\documentclass[a4paper,12pt]{article}
\usepackage[top=3cm, bottom=3cm, left=3cm, right=3cm]{geometry}
\usepackage{amsmath}
\usepackage{amssymb}
\usepackage{graphicx}
\usepackage{breqn}
\usepackage{multirow}
\usepackage{textcomp}

\DeclareMathOperator{\rrr}{r}
\DeclareMathOperator*{\argmax}{arg\,max}
\DeclareMathOperator{\cca}{CCA}
\DeclareMathOperator{\pcca}{PCCA}
\DeclareMathOperator{\kpcca}{KPCCA}

\DeclareMathOperator{\abs}{abs}
\newcommand\independent{\protect\mathpalette{\protect\independenT}{\perp}}
\def\independenT#1#2{\mathrel{\rlap{$#1#2$}\mkern2mu{#1#2}}}

\begin{Huge}
\title{Robust Non-linear Wiener-Granger Causality For Large High-dimensional Data}
\end{Huge}
\author{Mehrdad Jafari-Mamaghani$^{1,2}$}
\date{April 2014}
\begin{document} 
\maketitle
\begin{flushleft}
\begin{enumerate}
\item Division of Mathematical Statistics, Department of Mathematics, Stockholm University, Stockholm, Sweden
\item Center for Biosciences, Department of Biosciences and Nutrition, Karolinska Institutet, Huddinge, Sweden
\end{enumerate}
\end{flushleft}
\begin{abstract}
Wiener-Granger causality is a widely used framework of causal analysis for temporally resolved events.
We introduce a new measure of Wiener-Granger causality based on kernelization of partial canonical correlation analysis 
with specific advantages in the context of large high-dimensional data.
The introduced measure is able to detect non-linear and non-monotonous signals, 
is designed to be immune to noise, 
and offers tunability in terms of computational complexity in its estimations.
Furthermore, we show that, under specified conditions, the introduced measure 
can be regarded as an estimate of conditional mutual information (transfer entropy).
The functionality of this measure is assessed using comparative simulations where it outperforms other existing methods. 
The paper is concluded with an application to climatological data.
\end{abstract}
\section{Introduction}
\indent

Big data is a frequent topic in contemporary discourse on applications of statistics \cite{bigdata}; 
a topic that is likely to sustain its weight with diminishing costs of data storage and increasing accessibility to digital data. 
Here, a challenge, or an opportunity, is to move beyond identifying correlations to discovering causations 
using probability theory and statistics \cite{numbersense}. 
In statistics, frameworks of causal analysis include the Neyman-Rubin models of causal inference, Bayesian networks, and models of Wiener-Granger causality.
The first class of models are designed for experimental static data and thus, inappropriate for large observational data.\\ \indent
Bayesian networks are based on a conceptualization of causal learning on static data using probabilistic graphical models with 
algorithms operating on estimated (or a priori known) probability densities \cite{pearl}.
Bayesian networks can be extended to time-resolved domains using \textit{dynamic} Bayesian networks (DBN), 
a generalization of hidden Markov models \cite{MLPP}.
However, prior knowledge of probability densities is an uncommon occurrence in observational data and density estimation 
is a non-trivial task in high-dimensional settings.\\ \indent
Wiener-Granger causality operates on stochastic processes and measures causality by using the relative reduction in prediction error.
As Wiener-Granger causality has been proven to outperform DBN in large data sets \cite{DBN_GC},
and as it additionally offers frequency domain decomposition \cite{DBN_GC,geweke82}, 
we argue that Wiener-Granger causality is the most suitable choice for modeling causality in the context of (temporally resolved) 
large high-dimensional data. \\ \indent
Wiener-Granger causality (WGC), first conceived by Wiener \cite{wiener}, 
coined and parametrized by Granger \cite{granger69}, 
and developed further by Sims \cite{sims} and Geweke \cite{geweke84},
has led to a rich repertoire of analytical methods, 
conceptual spin-offs, and applications in social and natural sciences.
Dominant fields of applications of WGC include econometrics \cite{iranGC,casinoGC,miltaryGC}, 
neurophysiology \cite{stephanrev,depressGC,battcon11},
climatology \cite{clima1,clima2,clima3}, 
and most recently in cell biology \cite{pone}.
However, it should be noted that although WGC is not synonymous with causality in a philosophical sense, 
it offers a solid framework to quantitatively measure and verify a specific type of causality.\\ \indent
WGC conceptualizes the event where the cause temporally precedes the effect, and 
where the embedded information in the cause about the effect is unique \cite{granger69}.
Employing the nomenclature of probability theory and omitting vector notations without any loss of generality, 
under the null hypothesis, given $k$ lags and the temporally resolved random continuous vectors $A, B$ 
and the set of all other random vectors $C$ in any arbitrary system (or probability space $\Omega$), $\{B\}$ 
does not Wiener-Granger cause $\{A\}$, if:
\begin{align} \label{h00}
H_0:A_{t} \independent \{B_{t-1},\dots,B_{t-k}\} \vert \{A_{t-1},\dots,A_{t-k},C_{t-1},\dots,C_{t-k}\},
\end{align}
where $\independent$ denotes probabilistic independence and where we have assumed stochastic processes that are time-invariant/stationary.
For the sake of simplicity, we implement the following substitutions in the remainder of this study: 
$X=A_{t}$, 
$Y=\{ B \}_{t-1}^{t-k} $, and
$Z=\{ A, C \}_{t-1}^{t-k}$.
Accordingly, \eqref{h00} can be reformulated as: 
\begin{align} \label{H0}
H_0: X \independent Y \vert Z,
\end{align}
under which \cite{flomo}:
\begin{align}\label{Hf}
f(X|Z)=f(X|YZ).
\end{align}
Under the alternate hypothesis $H_1:X \not\independent Y \vert Z$, where $f(X|Z)\neq f(X|YZ)$ we say that $Y$ (Wiener-Granger) causes $X$ given $Z$.\\ \indent
Moving beyond the preliminaries, the next task is to test the hypothesis in \eqref{H0}. 
The method proposed by Granger to test this hypothesis is based on linear models and a corresponding variance-based test statistic \cite{granger69}.
More specifically, using linear regression to model the probability density functions in \eqref{Hf}, 
the hypothesis in \eqref{H0} can be tested using the F-distributed Granger-Sargent test based on restricted and unrestricted residual sums of squares.
Succeeding models have extended this concept to the multivariate setting using the \textit{generalized} variance \cite{geweke82,genvarGC} 
and the \textit{total} variance \cite{totvarGC}.
Other linear models have utilized feature selection techniques such as the Lasso in high-dimensional applications of WGC \cite{lassoGC1,lassoGC2}.
Naturally, the major drawback of linear models is their insensitivity to non-linear relationships, a common occurrence in empirical data. 
This shortcoming is often circumvented by the usage of non-linear (but parametric) models of WGC. 
To name a few, non-linear Fourier and wavelet transformations \cite{fourierGC,waveletGC}, 
radial basis functions \cite{ancona04,marina06}, 
copula density estimations \cite{copulaGC}, 
and additive models \cite{additiveGC}, 
have all been employed in non-linear analysis of WGC.
Although the choice of model can be a non-trivial issue, the primary disadvantage of non-linear models 
lies in the risk of overfitting, which usually leads to retrieval of spurious causations.\\ \indent
Another class of models used in analysis of WGC are based on the information theoretical approach and hence non-parametric.
Where parametric models are based on the covariance, non-parametric models employ the Shannon entropy \cite{shannon}
to quantify the deviation between the probability density functions in \eqref{Hf}. 
The most well-known measure in this context is the \textit{transfer entropy} \cite{transferentropy}, 
equivalent to the conditional mutual information \cite{hlava07,seg12,mjm,boston}.
Interestingly, there is an equivalence between transfer entropy and the linear generalized variance test of WGC for Gaussian variables \cite{barnett,mdpi}. 
Although non-parametric models offer an attractive framework for analysis of WGC, it has been observed that estimates of 
information theoretical measures suffer from insensitivity to non-linear relationships \cite{kraskov}.
This issue is potentially exacerbated in the context of high-dimensional data where estimation of multivariate probability densities,
needed to estimate information theoretical measures, 
is a non-trivial task and requires a fair amount of supervision and tuning. \\ \indent
Canonical correlation analysis (CCA) \cite{cca}, another statistical tool employed in analysis of WGC,
can be seen as an extension of multiple regression to multivariate regression, i.e. reduced rank regression.
In fact the solutions to reduced rank regression can be estimated via CCA \cite{rrrcca,esl}.
Moreover, it has been shown that the solutions from CCA are equivalent to those of orthonormalized partial least squares \cite{CCAOPLS}.
Given two multivariate sets of data, CCA finds linear projections of each set such that the between-projection correlations are maximized. 
Similar to principal component analysis, the projections are ordered based on the eigenvalues of covariance matrices; 
a property that is of particular interest in the presence of elevated levels of noise \cite{GCfMRI?}.
In the context of WGC, CCA has been employed in order to overcome singularities resulting from high-dimensionality of data \cite{ccaeco}, and 
partial CCA (PCCA) has been employed in neural connectivity analysis \cite{wuPCCA}, and studies of optical flows in videos \cite{cflow}.\\ \indent
In this paper, we will introduce a new measure of Wiener-Granger causality based on kernel PCCA. 
More specifically, we will derive a measure of WGC based on kernel partial canonical correlation analysis (KPCCA) in the reproducing 
kernel Hilbert spaces (RKHS). 
The issues of overfitting and computational complexity are tackled by penalization and parsimonious matrix decomposition, respectively. 
In addition, we will show that under certain conditions, the measure based on KPCCA can be regarded as an estimate of transfer entropy 
(conditional mutual information), 
a particularly appealing property since estimation of transfer entropy in high-dimensional settings is a non-trivial and expensive task \cite{aste}.
We will also discuss the similarities and differences between the measure in this paper and closely related measures in the context of WGC.\\ \indent
In the following, 
Section 2 introduces the methods, 
Section 3 presents data simulations, climatological data, and corresponding results. 
Lastly, Section 4 closes the paper with the concluding remarks.
\section{Methods}\label{meths}
\indent

In this section we will build up the framework needed to introduce KPCCA.
We will adopt the same variables as those used in the Introduction and follow the hypothesis in \eqref{H0} unless otherwise stated. 
Furthermore, without any loss of generality, we assume that all multivariate random vectors in the remainder of 
this paper are standard Gaussians with $N$ observations, mean zero and standard deviation 1, 
and denote the dimension of each random vector by $d_X$, $d_Y$ and so on.
\subsection{Canonical correlations}
\subsubsection{Correlation analysis}
\indent

The product-moment correlation coefficient between any two random variables $A$ and $B$ is defined by:
\begin{align}\label{pcc1} 
\rho_{A,B} &= \frac{\Sigma_{AB}}{\sigma_A \sigma_B} 
\end{align}
where $\Sigma_{AB}$ represents the covariance of $A$ and $B$, and $\sigma_{A}$ the standard deviation of $A$. 
Following the earlier specification of centered random variables, $\rho_{A,B}$ can be reformulated as:
\begin{align}\label{pcc2}
\rho_{A,B} &= \frac{A'B}{\sqrt{A'A}\sqrt{B'B}}\\
&=\frac{\langle A, B \rangle}{{ \langle A, A\rangle^{1/2} }{ \langle B,B\rangle^{1/2} } } \label{innerp}
\end{align}
where $A'$ denotes the transpose of $A$ in \eqref{pcc2}, and $\langle A,A \rangle$ is called the \textit{inner product} of $A$ in \eqref{innerp}.
The non-parametric variety of $\rho$ is the Spearman's rank correlation coefficient $\rho^{(\rrr)}$ and is simply based on the ranks of $A$ and $B$.
Denoting the ranks of $A$ and $B$ as $A^{(\rrr)}$ and $B^{(\rrr)}$, respectively, 
the value of $\rho^{(\rrr)}$ is determined by plugging in $A^{(\rrr)}$ and $B^{(\rrr)}$ in \eqref{pcc1}.\\

\subsubsection{Canonical correlation analysis}
\indent

Formulated in 1936 by Hotelling \cite{cca},
canonical correlation analysis (CCA) extends the concept in the previous section to two (or more) \textit{sets} of variables.
Re-employing the random vectors $X$ and $Y$, CCA is designed to find linear projections of $X$ and $Y$ subject to the maximization of 
correlations between these projections. 
The linear projections yielded by CCA are closely linked to the linear projections in principal component analysis (PCA), 
and the condition of between-set correlation maximization can be regarded as a generalization of the solution in partial least squares (PLS) \cite{KICA}.\\\indent
Let $\alpha$ and $\beta$ denote linear projections of the multivariate vectors $X$ and $Y$, respectively. 
The \textit{canonical correlations} of $X$ and $Y$ are defined by:
\begin{align}
\rho^{(\cca)}_{X,Y} & = \smash{\displaystyle\argmax_{\alpha,\beta}} \quad \rho_{X \alpha, Y \beta} \nonumber \\
&=\smash{\displaystyle\argmax_{\alpha,\beta}} \quad \frac{\langle X \alpha,  Y \beta \rangle }
{\langle X \alpha,   X \alpha \rangle^{1/2} \langle Y \beta,  Y \beta \rangle^{1/2} }. \label{cca}
\end{align}
Taking the derivatives of \eqref{cca} with respect to $\alpha$ and $\beta$ yields:
\begin{align}
\frac{\partial}{\partial \alpha} \rho^{(\cca)}_{X,Y} &= 
\langle X,Y \rangle \beta - \frac{ \langle X \alpha , Y \beta \rangle }{\langle X \alpha , X \alpha \rangle } \langle X, X \rangle \alpha  \nonumber \\
\frac{\partial}{\partial \beta} \rho^{(\cca)}_{X,Y} &= 
\langle Y,X \rangle \alpha - \frac{ \langle X \alpha , Y \beta \rangle }{\langle Y \beta , Y \beta \rangle } \langle Y, Y \rangle \beta  \nonumber.
\end{align}
\indent
Thus, by setting the derivatives to zero and letting $\langle X \alpha , X \alpha \rangle = \langle Y \beta, Y \beta \rangle=1$,
we arrive at the following generalized eigenvalue problem, using which, one can estimate $\alpha$ and $\beta$:
\begin{align} \label{sols1}
\begin{pmatrix}   0 & \langle X, Y\rangle \\   \langle Y, X\rangle & 0  \end{pmatrix}
\begin{pmatrix}   \alpha \\   \beta  \end{pmatrix}= \rho^{(\cca)}_{X,Y}
\begin{pmatrix}   \langle X, X\rangle & 0\\   0 & \langle Y, Y\rangle  \end{pmatrix}
\begin{pmatrix}   \alpha \\   \beta  \end{pmatrix}.
\end{align}
\indent
The eigenvalues of the system in \eqref{sols1} lead to $d=\min(d_X,d_Y)$ canonical correlations. 
The solution above can be extended to more than two sets of variables as demonstrated in \cite{ketten,KCCA}.\\

\subsubsection{Partial canonical correlation analysis}
\indent

The first reformulation of CCA to include conditionality was by Rao in 1969 where the canonical correlations of $X$ and $Y$ 
conditioned on a third vector $Z$ are retrieved by removing the influence of $Z$ using regression techniques \cite{pcca}. 
Here we pursue a similar concept to introduce partial canonical correlation analysis (PCCA) based on \textit{partial} covariances.\\\indent
Let $\rho^{(\pcca)}_{X,Y \vert Z}$ denote the partial canonical correlations of $X$ and $Y$ given the information in $Z$: 
\begin{align}
\rho^{(\pcca)}_{X,Y \vert Z} &=\smash{\displaystyle\argmax_{\alpha,\beta}} \quad \frac{\langle  (X \vert Z) \alpha, (Y \vert Z)\beta \rangle }
{\langle (X \vert Z) \alpha,(X \vert Z) \alpha \rangle^{1/2} \langle(Y \vert Z) \beta,(Y \vert Z)\beta \rangle^{1/2} } \nonumber
\end{align}
where $X | Z$ denotes the random vector $X$ with discarded influence of $Z$.
Using a scheme similar to the previous section, the partial canonical correlations are determined by the following generalized eigenvalue problem:
\begin{dmath} \label{sols_pcca}
\begin{pmatrix}   0 & \langle X, Y\rangle - \langle X, Z\rangle \langle Z,Z\rangle^{-1} \langle Z,Y\rangle \\   
\langle Y, X\rangle - \langle Y, Z\rangle \langle Z,Z\rangle^{-1} \langle Z,X\rangle & 0  \end{pmatrix}
\begin{pmatrix}   \alpha \\   \beta  \end{pmatrix}
= \rho^{(\pcca)}_{X,Y\vert Z}
\begin{pmatrix}   \langle X, X\rangle - \langle X, Z\rangle \langle Z,Z\rangle^{-1} \langle Z,X\rangle & 0\\   
0 & \langle Y, Y\rangle - \langle Y, Z\rangle \langle Z,Z\rangle^{-1} \langle Z,Y\rangle \end{pmatrix}
\begin{pmatrix}   \alpha \\   \beta  \end{pmatrix}
\end{dmath}
where the non-zero elements in the quadratic matrices are partial covariances.

\subsubsection{Kernel partial canonical correlation analysis}
\indent

Kernel methods have had a substantial impact in statistical learning techniques in the past decade \cite{MLPP,esl,kernelbook}.
Kernel methods accommodate the analysis of non-linear phenomena using linear techniques by mapping data 
from an input space into a higher-dimensional feature space $\mathcal{F}$:
\begin{align}\nonumber
\Xi : X_1, \ldots ,X_{d_X} \longmapsto \Xi(X_1),\ldots,\Xi(X_{d_X}).
\end{align}
\indent
In the context of large data sets, \textit{Mercer kernels} are of particular utility.
Let $X$ be defined on the space $\mathcal{X}$, and let $\kappa(X_i,X_j)$ be a function from $\mathcal{X} \times \mathcal{X}$ to $\mathbb{R}$.
If the matrix $K$, defined element-wise as $K_{ij}=\kappa(X_i,X_j)$, is a positive semidefinite matrix, 
the function $\kappa(X_i,X_j)$ is a \textit{Mercer kernel}, 
and the matrix $K$ a \textit{Gram matrix}.
Under these conditions there is a space $\mathcal{F}$ and a map $\Xi$ such that $\kappa(X_i,X_j)$ is the inner product in $\mathcal{F}$
between the images $\Xi(X_i)$ and $\Xi(X_j)$:
\begin{align}\nonumber
\langle \Xi(X),\Xi(X) \rangle = \kappa(X,X) = K_{X,X}.
\end{align}
A continuous kernel qualifying as a Mercer kernel is the Gaussian kernel: $\Phi(X)=\exp(-X^2 /2)/\sqrt{2\pi}$, which in addition 
associates $\kappa(\cdot,\cdot)$ with a \textit{reproducing kernel Hilbert space} (RKHS) \cite{kernelbook}. 
The appealing aspect of the theory above lies within the properties of the Gram matrix $K$: 
under the Mercer conditions, $K$ can conveniently be factorized to lower-dimensional matrices 
to counterbalance computational complexity (see below). \\\indent
The usage of kernel methods in analysis of WGC is not an uncommon practice. 
A potent approach is presented in \cite{KGC} where using statistical tests, 
the Gram matrices used in non-linear analysis of WGC are pruned to circumvent overfitting. 
More closely related to the topic of this paper, a kernelization of CCA in analysis of WGC is presented in \cite{KCCAwu}. 
However, the presented framework here is not immune to the likely issues of overfitting and overcomplexity (cf. the sections below).\\ \indent
Assuming centered Gram matrices in the remainder, 
the kernel partial canonical correlations of $X$ and $Y$ given the information in $Z$ are defined by:
\begin{align}\label{kpcca}
\rho^{(\kpcca)}_{X,Y \vert Z} &=\smash{\displaystyle\argmax_{\gamma,\delta}} \quad \frac{\langle \Phi(X\vert Z) \gamma, \Phi( Y \vert Z)\delta \rangle }
{\langle \Phi(X\vert Z) \gamma,\Phi(X\vert Z) \gamma \rangle^{1/2} \langle \Phi(Y\vert Z) \delta,\Phi(Y\vert Z) \delta \rangle^{1/2} }
\end{align}
where $\Phi(X\vert Z)$ denotes the image of the random vector $X$ in the feature space $\mathcal{F}$ with removed influence of $Z$.\\ \indent
The kernelization of PCCA as conducted above can lead to two issues in real-world applications.
Firstly, the naive kernelization in \eqref{kpcca} can yield biased canonical correlations. 
That is, due to the potential risk of overfitting, the kernel-defined spaces spanned by $\langle \Phi(X\vert Z) ,\Phi(X\vert Z) \rangle$ 
and $\langle \Phi(Y\vert Z) ,\Phi(Y\vert Z) \rangle$ are likely to
be oriented in identical directions and result in perfect canonical correlations \cite{KCCA}.
Secondly, the parametrization in \eqref{kpcca} requires the evaluation of full-rank Gram matrices, 
which is likely to lead to a prohibitive computational expenditure for large data samples.\\ \indent
In the following, the two issues of overfitting and computational efficiency are addressed using regularization and incomplete 
Cholesky decomposition, respectively.
\subsubsection{Regularization}
\indent

The issue of overfitting, leading to falsely high correlations, can be circumvented by regularization.
We use the partial least squares norm penalization with regularization parameter $\zeta$ to reformulate $\rho^{(\kpcca)}_{X,Y \vert Z}$:\\
\begin{align}\label{kpcca2}
\rho^{(\kpcca)}_{X,Y \vert Z} &=\smash{\displaystyle\argmax_{\gamma,\delta}} \quad \frac{\langle \Phi(X\vert Z) \gamma, \Phi( Y \vert Z)\delta \rangle }
{  \Psi(\Phi(X\vert Z) \gamma,\zeta) ^{1/2}   \Psi(\Phi(Y\vert Z) \delta,\zeta)  ^{1/2} }
\end{align}
where
\begin{align}
\Psi(\Phi(X\vert Z) \gamma,\zeta)&=\langle \Phi(X\vert Z) \gamma,\Phi(X\vert Z) \gamma \rangle + 
\zeta \| \langle \gamma,\Phi(X\vert Z) \gamma \rangle \|^2 \label{deno1} , \\
\Psi(\Phi(Y\vert Z) \delta,\zeta)&=\langle \Phi(Y\vert Z) \delta,\Phi(Y\vert Z) \delta \rangle + 
\zeta \| \langle \delta,\Phi(Y\vert Z) \delta \rangle \|^2. \label{deno2}
\end{align}
\indent

By expanding the terms in \eqref{deno1} and \eqref{deno2}
up to the second order in $\zeta$ and deriving the derivatives of \eqref{kpcca2}, we arrive at the following generalized
eigenvalue problem:
\begin{align} \label{sols_kpcca2}
\begin{pmatrix}   0 & K_{X,Y\vert Z} \\  K_{Y,X\vert Z}  & 0  \end{pmatrix}
\begin{pmatrix}   \gamma \\   \delta  \end{pmatrix}= \rho^{(\kpcca)}_{X,Y \vert Z}
\begin{pmatrix}   K_{X ,X\vert Z}+\zeta I & 0\\   0 & K_{Y,Y\vert Z}+\zeta I  \end{pmatrix}
\begin{pmatrix}   \gamma \\   \delta  \end{pmatrix}
\end{align}
where $I$ is the identity matrix and
\begin{align}\nonumber
K_{X ,Y \vert Z}=\langle \Phi(X), \Phi(Y) \rangle - \langle \Phi(X), \Phi(Z) \rangle \langle \Phi(Z),\Phi(Z)\rangle^{-1} \langle \Phi(Z),\Phi(Y)\rangle.
\end{align}
The solution obtained by solving \eqref{sols_kpcca2} is similar to the solution yielded by the \textit{Canonical Ridge} \cite{cflow,CCAridge}.
\subsubsection{Incomplete Cholesky decomposition}\label{ICD} 
\indent

Given a positive semidefinite $d \times d$ matrix $K$, Cholesky decomposition finds a factorization of $K$ such that $K=GG'$, 
where $G$ is a lower triangular $d \times d$ matrix, and $G'$ denotes the transpose of $G$. 
Our aim is to approximate $G$ using a low-rank $d \times c$ matrix $\widetilde{G}$ where $c \ll d$, 
and the difference $K-\widetilde{G}\widetilde{G}'$ has a norm less than a given threshold $\tau$.
This aim is achieved by incomplete Cholesky decomposition \cite{ICD}.
Whereas complete Cholesky decomposition employs all pivots in factorization, incomplete Cholesky decomposition skips those pivots 
that are below a given level.
In fact, incomplete Cholesky decomposition is the dual implementation of partial Gram-Schmidt orthogonaliztion \cite{KCCA}.
The complete algorithm for incomplete Cholesky decomposition can be found in \cite{KCCA}.
\subsection{Measures of causality}
\indent

Using the same random vectors as in \eqref{H0}, 
and the definition of partial canonical correlations above, we define the measure \textit{Canonical Wiener-Granger Causality} (CC) as:
\begin{align}\label{WGC}
\mathcal{C}_{Y \rightarrow X \vert Z } := -\frac{1}{2} \ln \prod_{i=1}^{d} \left[ 1 - \left( \rho_{X,Y \vert Z}^{(\pcca)} (i) \right) ^2 \right]
\end{align}
where $\rho_{X,Y \vert Z}^{(\pcca)} (i)$ denotes the $i$th partial canonical correlation, and $d=\min(d_X,d_Y)$.
As it turns out, when $d=d_X=d_Y$ for Gaussian variables, the measure in \eqref{WGC} is equivalent 
to the transfer entropy of $X$ and $Y$ given $Z$:
\begin{align}\label{WGC_eq}
\mathcal{C}_{Y \rightarrow X \vert Z }=\mathcal{T}_{Y \rightarrow X \vert Z }.
\end{align}
A proof of \eqref{WGC_eq} is given in \ref{CWGCTE}.
As a consequence, under the specified conditions, the maximum likelihood estimator of \eqref{WGC} 
follows a $\chi^2$ distribution under the null hypothesis (asymptotically for large samples).
In any other case, tests of significance can be carried out using permutation resampling.\\ \indent
Similarly, we define the measure \textit{Kernel Canonical Wiener-Granger Causality} (KCC) as:
\begin{align}\label{kWGC}
\mathcal{C}_{Y \rightarrow X \vert Z }^{\mathcal{F}} := -\frac{1}{2} \ln \prod_{i=1}^{d^{(\mathcal{F})}} 
\left[ 1 -  \left( \rho_{X,Y \vert Z}^{(\kpcca)} (i) \right) ^2 \right].
\end{align}\indent
Accordingly, given Gaussian distributed images of input data in the higher-dimensional feature space $\mathcal{F}$, 
where $d^{(\mathcal{F})}=d^{(\mathcal{F}_X)}=d^{(\mathcal{F}_Y)}$, we can show that
$\mathcal{C}_{Y \rightarrow X \vert Z }^{\mathcal{F}} =\mathcal{T}_{Y \rightarrow X \vert Z }^{\mathcal{F}}$,
where the term on the right hand side denotes the transfer entropy of $\Phi(Y)$ to $\Phi(X)$ given $\Phi(Z)$ in the feature space $\mathcal{F}$.
The proof follows by analogy from \ref{CWGCTE}.\\ \indent
Although the equality in \eqref{WGC_eq} provides a useful estimate of transfer entropy using PCCA, 
the assumption of Gaussianity for observed processes is not always met in real-world applications. 
However, using the Gaussian kernel $\Phi(X)=\exp(-X^2 /2)/\sqrt{2\pi}$, we argue that:
\begin{align}\label{kWGC_eq}
\mathcal{C}_{Y \rightarrow X \vert Z }^{\mathcal{F}} =\mathcal{T}_{Y \rightarrow X \vert Z }.
\end{align}
\indent
That is, the KCC based on images of data in the feature space $\mathcal{F}$ derived via Gaussian kernels, 
coincides with the transfer entropy of data in the input space. 
This result follows from the relation between kernel generalized variance and mutual information. 
Further remarks underlying the equality in \eqref{kWGC_eq} are given in \ref{KCWGCTE}.\\ \indent
It follows from \eqref{kWGC_eq} that:
\begin{align}
\mathcal{T}_{Y \rightarrow X \vert Z }^{\mathcal{F}}=\mathcal{T}_{Y \rightarrow X \vert Z },
\end{align}
leading to the conclusion that the transfer of information from $Y$ to $X$ given $Z$ is identical 
regardless of its evaluation in the input or the feature space.
The intuition here is underpinned by the fact that kernelization of data only serves to make information more accessible rather than altering its content.\\\indent
The utilities of KCC as a measure of causality can be grouped into:
\textit{i)} its innate properties, and \textit{ii)} its advantage over estimates of entropy-based measures.
Firstly, KCC is preferable to linear measures as it is capable of detecting non-linear causal signals.
Furthermore, as we shall see in applications to synthetic data, 
by using the eigenvalues of Gram matrices, KCC is founded only on the canonical information, disregarding spurious signals. 
This feature is highly relevant in the context of high-dimensional input spaces where the pervasiveness of noise may lead 
to difficulties in discovering the underlying relationships.
Additionally, by employing incomplete Cholesky decomposition, we are able to control the computational complexity of the 
main bottleneck in estimations of KCC. \\\indent
Secondly, KCC provides a superior alternative to estimates of differential/continuous transfer entropy based on kernel density 
estimation (KDE) \cite{silver86,wand95} as these estimates, in the context of estimating information theoretical measures, suffer from insensitivity to non-linear relationships \cite{kraskov,mjm}.
Moreover, KDE-based estimates need supervision in terms of bandwidth selection and demand substantial computational resources in 
high-dimensional settings.\\\indent
In the following, we will assess the performance of CC and KCC using synthetic and real-world data.
\section{Results}\label{res}
\subsection{Synthetic data}\label{sims}
\indent

\begin{figure}[!ht]
\centering
\includegraphics[angle=0,width=15.5cm,clip=true,trim=150 50 120 50]{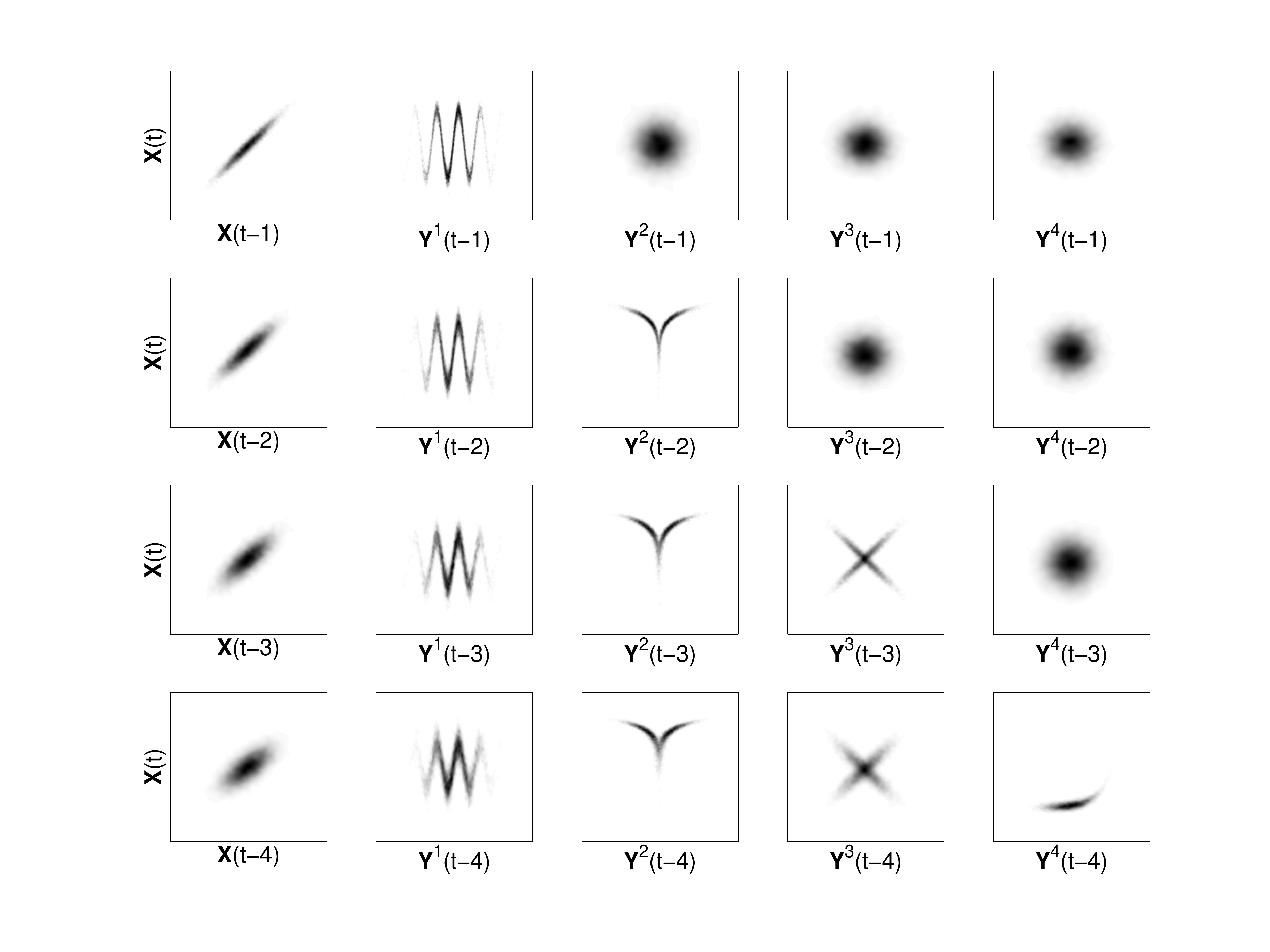}%
\caption{\textit{Density plots of $X_t$ against all lags of other variables.
The system exhibits different types of non-linear and/or non-monotonous relationships. 
All other inter-variable relationships excluded from this figure display white-noise processes (see top right panels of this figure).}}\label{scats}
\end{figure}

\begin{center}
\begin{figure}[!ht]
\centering
\includegraphics[angle=0,width=15.5cm,clip=true,trim=45 50 120 0]{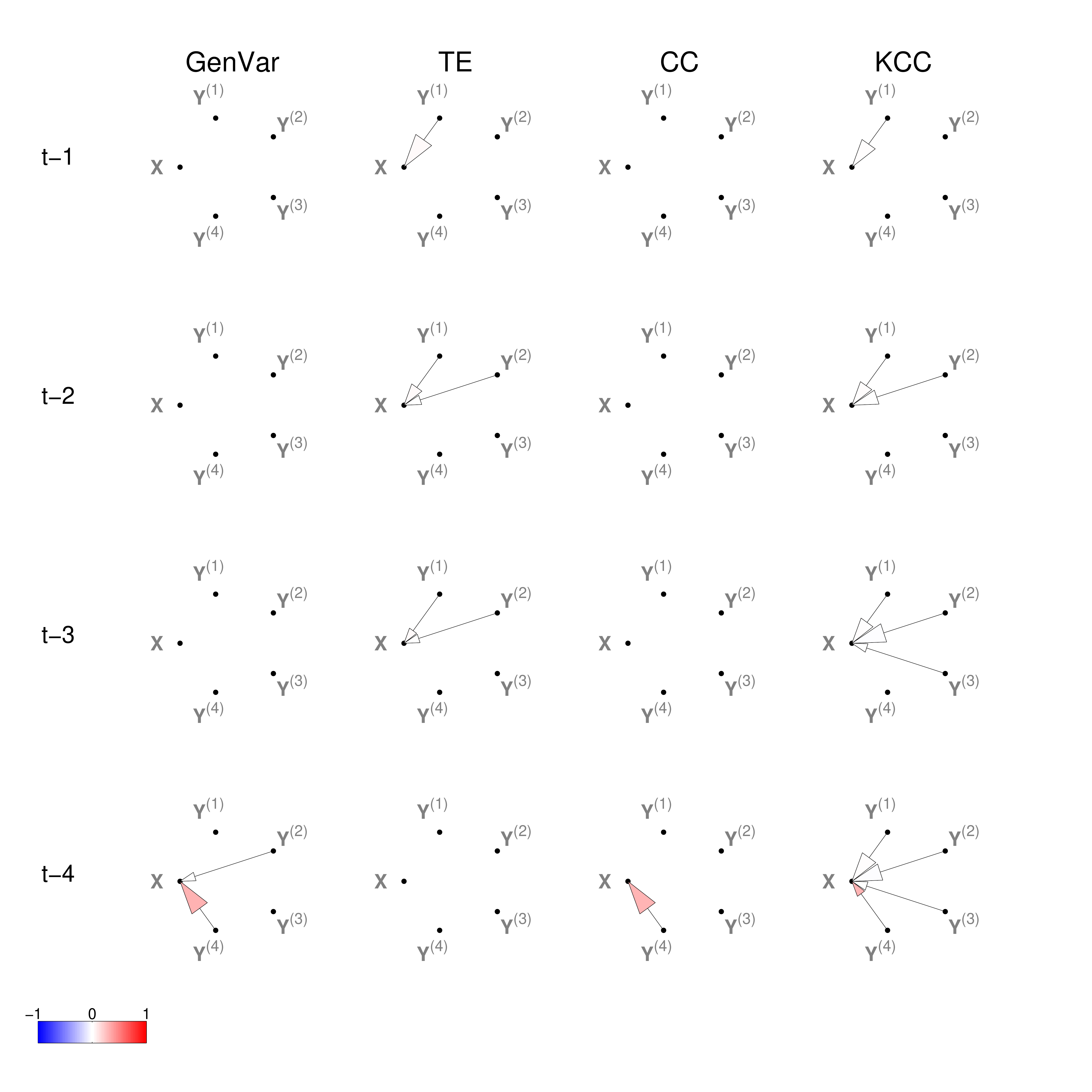}%
\caption{\textit{Results from the simulations in \ref{sims} displayed as networks 
where a critical value of $\alpha=0.001$ has been set to rule out non-causality.
In the simulations, KCC outperforms the other methods as it successfully detects the planted causal signals.
Both GenVar (generalized variance) and CC detect the monotonous exponential signal.
Transfer entropy (TE) detects the causal signals from $Y^{(1)}$ and $Y^{(2)}$ to $X$ in up to $k=3$ lags but fails to discover the remaining signals
in higher dimensions.
The size of the arrows displays the relative distance of the empirical causal measure from the 99th percentile of the distributions under the null hypotheses.
The color of the arrow displays an aggregate score of rank correlations 
(the rank correlation of average ranks of the two vectors).}}\label{nets}
\end{figure}
\end{center}

\begin{figure}[!ht]
\centering
\includegraphics[angle=90,width=13cm,clip=true,trim=50 10 150 20]{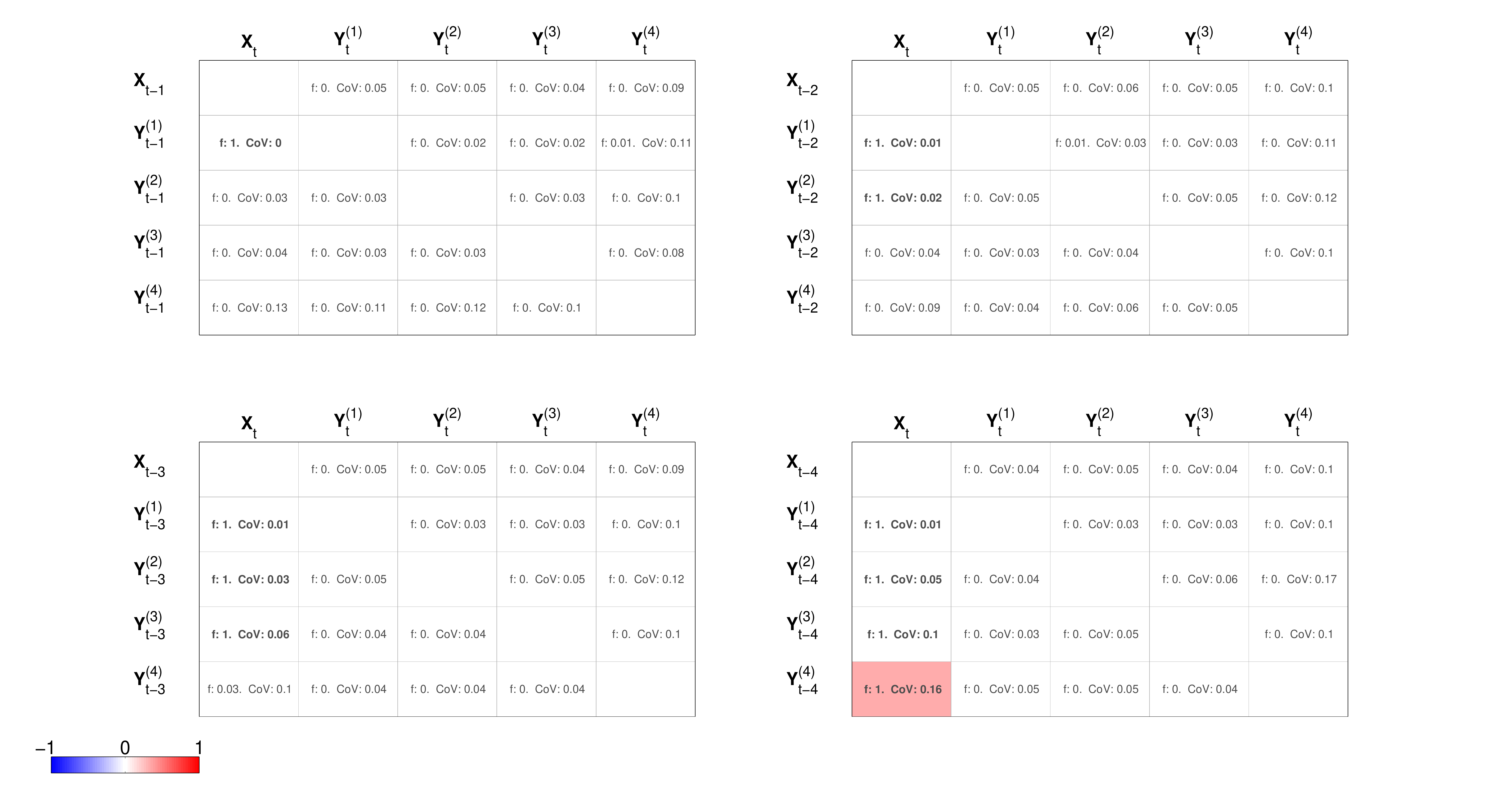}%
\caption{\textit{Results on the performance of KCC on 100 realizations of the system presented in \ref{sims}. 
Causal signals go from row to column, $f$ denotes the discovery rate of a causal signal, 
and $CoV$ denotes the coefficient of variation of the KCC scores. The color-coding is based on aggregate score of rank correlations
(the rank correlation of average ranks of the two vectors).}}\label{KCCdet} 
\end{figure}

To assess the functionality of the proposed measures of causality, CC and KCC, we will use the system in \eqref{sim} to simulate 
synthetic multivariate time-series exhibiting non-monotonous causal signals. 
We define the 20-dimensional multivariate vectors ${X}, {Y}^{(1)}, {Y}^{(2)}, {Y}^{(3)}$ and ${Y}^{(4)}$ as:
\begin{align}\label{sim}
\left.  \begin{aligned}
&  \left\{  \begin{array}{l l l}
  {X}_t = {s} ,& {s} \sim \mathcal{N}({0},{I})\\
  {X}_{t-i} = {s} + 0.1 \cdot i \cdot \epsilon  ,& i=1 ,\ldots, k, &  \epsilon \sim \mathcal{N}({0},{I})
  \end{array}   \right. \\
&  \left\{  \begin{array}{l l}
  {Y}_{t}^{(1)} = \epsilon &  \epsilon \sim \mathcal{N}({0},{I})\\
  {Y}_{t-i}^{(1)} = \sin(5{X}_t )+ 0.1 \cdot i \cdot \epsilon ,& i=1 ,\ldots, k \\
  \end{array}   \right. \\
&  \left\{  \begin{array}{l l}
  {Y}_{t-i}^{(2)} = \epsilon , & i=0,1\\
  {Y}_{t-j}^{(2)} = \log[\abs({X}_t + 0.25 \cdot (j-1) \cdot \epsilon )] ,& j=2,\dots,k \\
  \end{array}   \right. \\
&  \left\{  \begin{array}{l l l}
  {Y}_{t-i}^{(3)} = \epsilon, & i=0,\ldots,2\\
  {Y}_{t-j}^{(3)} = (-1)^{\omega}\cdot {X}_t + 0.2 \cdot(j-2)\cdot   \epsilon , & j=3,\ldots,k , & \omega \sim \mathcal{U}(1,2) \\
   \end{array}   \right. \\
&  \left\{  \begin{array}{l l}
  {Y}_{t-i}^{(4)} = \epsilon , & i=0,\ldots,3\\
  {Y}_{t-j}^{(4)} = \exp({X}_t)+ 1.3 \cdot (j-3) \cdot \epsilon, & j=4,\ldots,k\\
  \end{array}   \right.
\end{aligned}   \right.
\end{align}
where $\mathcal{N}({0},{I})$ denotes a multivariate standard Gaussian, 
and $\mathcal{U}(a,b)$ is a discrete uniform distribution.
Density plots in Figure \ref{scats} provide a display of the non-linear relationships in \eqref{sim}.
In our simulations we have used $k=4$ lags in 10,000 realizations of the system above.
The results, as a comparison between CC, KCC, GenVar (the generalized variance test, see \eqref{gewekegenvar}),
and transfer entropy/conditional mutual information as estimated in \cite{mjm},
using 1000 permutation resamplings, are presented in Figure \ref{nets}. 
In applications of KCC to the system above, the width of the Gaussian kernel has been set to $\sigma=1$ after standardization of data, 
the penalization parameter $\zeta=10^{-7}$,
and the incomplete decomposition of kernel matrices uses the threshold $\tau=10^{-6}$. 
In Figure \ref{nets}, the presence of an arrow from, e.g., $Y^{(1)}$ to $X$ at lag $k=2$, denotes the rejection of the hypothesis:
\begin{align*}
H_0 : X_t \independent Y^{(1)}_{t-1},Y^{(1)}_{t-2} \quad \vert \quad 
X_{t-1}, X_{t-2} , Y^{(2)}_{t-1}, Y^{(2)}_{t-2} , Y^{(3)}_{t-1}, Y^{(3)}_{t-2} ,Y^{(4)}_{t-1}, Y^{(4)}_{t-2}.
\end{align*}
\indent
From the networks in Figure \ref{nets}, it is apparent that GenVar and CC fail to discover the non-linear and non-monotonous relationships. 
However, generalized variance and CC do capture the exponential causal signal from ${Y}^{(4)}$ to ${X}$ due to its monotonicity.
Transfer entropy succeeds only in detecting the causal signals from $Y^{(1)}$ and $Y^{(2)}$ to $X$ for up to $k=3$ lags 
and thus, fails to find the remaining signals due to increased dimensionality and levels of noise.
In contrast, KCC successfully discovers all the planted relationships absent of any spurious causations.\\ \indent
Moreover, to extend the scope of analysis based on the performance of KCC, 100 realizations of the system in \eqref{sim} were performed to outline the 
robustness of KCC in terms of rate of discovery of causal links, and coefficients of variation of KCC scores (see Figure \ref{KCCdet}).
The results present unanimously low false discovery rates ($<3\%$), and uniformly consistent discovery of the planted signals. 
\subsection{Climatological data}
\indent

\begin{figure}[!ht]
\centering
\includegraphics[angle=90,width=11cm,clip=true,trim=60 30 135 30]{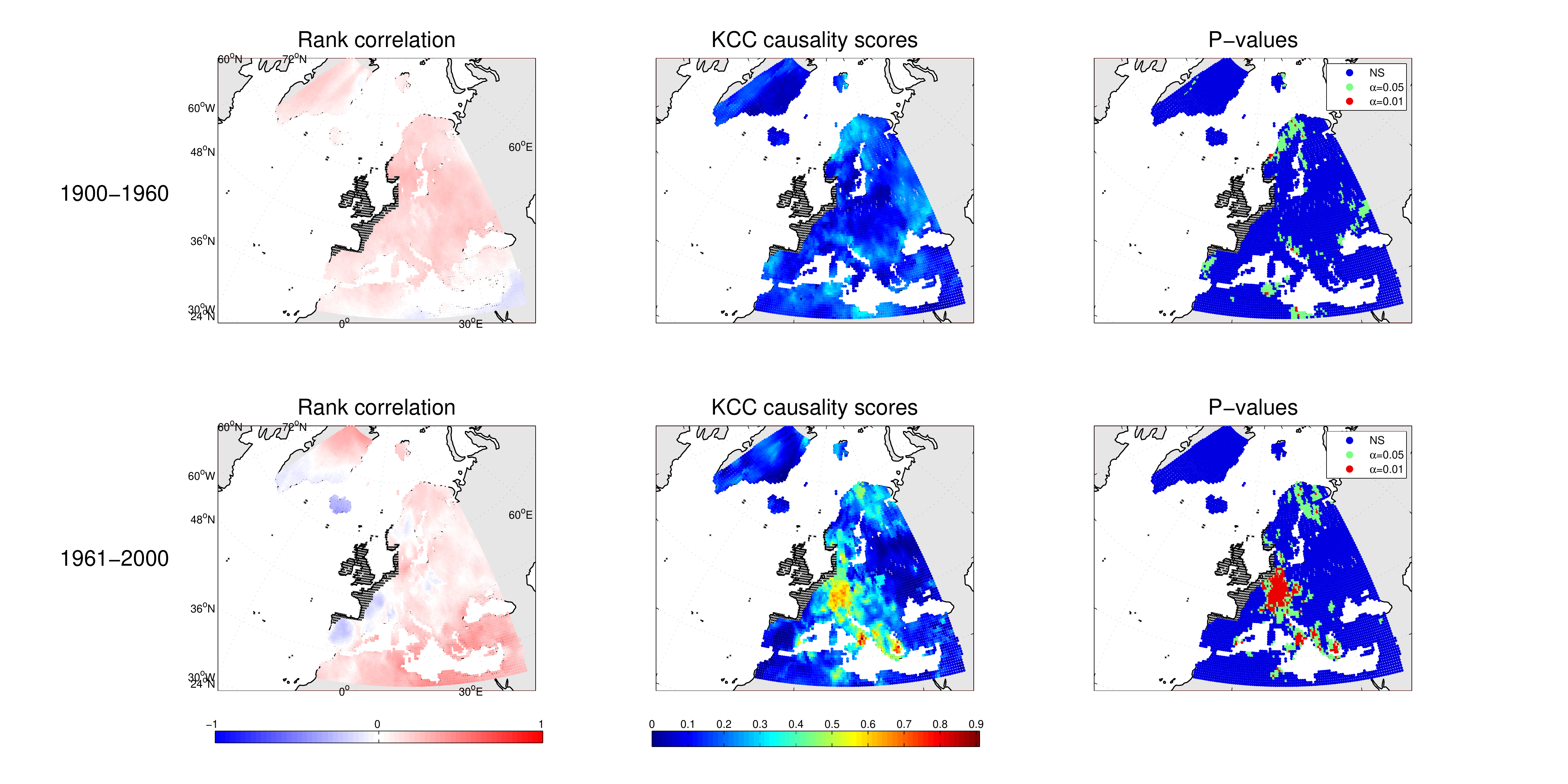}%
\caption{\textit{Analysis of Wiener-Granger causality based on the hypothesis in \eqref{climaH0}. 
The black-crossed is the North Atlantic region. 
The rank correlations present the correlations between temperature and precipitation in the North Atlantic region in March, 
and the temperature in rest of the continent in April. 
P-values are evaluated using permutation resamplings under the null hypothesis.
Maps from the M\_Map toolbox by Rich Pawlowicz.}}\label{eu1}
\end{figure}

To test the performance of KCC on real-world large high-dimensional data, an analysis of WGC is performed on gridded climatological records. 
It should be added that the aim here is not to draw any exclusive conclusions on climatological mechanisms, but to test the utility of KCC on types of 
large datasets that occur with increasing frequency.  \\ \indent
The data contains information on temperature and precipitation from December 1765 to November 2000, recorded in the 
North Atlantic/European region (80-30\textdegree N and 50\textdegree W-40\textdegree E) \cite{casty}; see Figure \ref{eu1}.
As climatological data reconstructed from proxies often suffers from relatively higher levels of noise than instrumental data \cite{bratt}, 
the analysis here is limited to the period 1900-2000 based on actual records.
Furthermore, the 20th century is divided into the two periods 1900-1960 and 1961-2000 
due to evidence of increased global warming effects in the latter period \cite{casty}.
Lastly, to ensure stationarity, i.e. avoid seasonal variations, we confine the analysis to a specific month. \\ \indent
In the applied analysis herein, we investigate whether the variation in temperature and precipitation in the North Atlantic regions of Europe 
(designation adapted from the European Environmental Agency, 
the black-crossed regions in Figure \ref{eu1}), 
is causal of temperature in other regions of Europe in the month April:
\begin{align}\label{climaH0}
H_0^{}: T^{(x)}_{April} \independent T^{(NA)}_{March}, P^{(NA)}_{March} \quad \vert \quad T^{(x)}_{March}, P^{(x)}_{March}.
\end{align}
In the hypothesis above $T$ denotes temperature, $P$ denotes precipitation, 
and the superscript $x$ denotes any region outside of the North Atlantic (NA) region 
(here, we regard every pixel in the gridded records outside of the NA region as an independent region). 
The results, based on tests of the hypothesis in \eqref{climaH0}, for the two periods 1900-1960 and 1961-2000 are displayed in Figure \ref{eu1}.
The application of KCC here uses the same parameters as in the previous section after data standardization.\\ \indent
It is seen that the variation of temperature and precipitation in the North Atlantic 
regions of Europe in March has a stronger causal effect on the April temperatures in the continental regions 
in the latter parts of the 20th century. 
In other words, in the period 1960-2000, temperature and precipitation in the North Atlantic regions in March contain unique information
about the temperatures of continental Europe in April;
a phenomenon that does not seem to be present in the earlier parts of the 20th century.\\ 

\section{Discussion}
\indent

Wiener-Granger causality is becoming a routine practice in causal analysis of temporally resolved observational data. 
In the context of large high-dimensional data, the issues of non-linearity, noise pervasiveness, and computational complexity 
present a series of challenges to classical frameworks of analysis of Wiener-Granger causality.
The aim of this paper has been to present an alternative approach to measure Wiener-Granger causality, which circumvents all of the three issues above.\\\indent
The proposed measure of Wiener-Granger causality (KCC) in this paper based on kernel partial canonical correlations and parsimonious matrix factorization 
offers the following advantages:
\textit{i}) as demonstrated by the simulations, KCC is sensitive to non-linear and non-monotonous signals absent of any display of spurious causalities,
\textit{ii}) as seen in the comparison between transfer entropy and KCC, 
by using the canonical information, KCC is more immune to noise and high-dimensionality,  
\textit{iii}) when measured using Mercer kernels, KCC offers appealing tunability in terms of computational complexity of its main estimation bottleneck
and is capable of operating considerably faster than similar kernel methods,
and \textit{iv}) most importantly, when measured using Gaussian kernels, KCC can be regarded as an estimate of transfer entropy/conditional 
mutual information.\\\indent
All of these properties offer clear advantages over currently existing measures in non-linear/non-parametric analysis of Wiener-Granger causality.
Given this background, we believe that KCC has the potential to become a standard measure 
as a result of its superior performance, computational feasibility, and straightforward implementation.
Future work on this topic includes closer investigation of KCC and its relation to other existing measures 
and more comprehensive applications to real-world data.
\section*{Acknowledgements}
\indent

The author wishes to thank John Hertz, Joanna Tyrcha and John G. Lock for insightful comments.
The author has been supported by 
the Swedish Research Council grant \# 340-2012-6011.

\appendix 
\section{Appendix} \label{app}
\subsection{Equivalence between CC and transfer entropy}\label{CWGCTE}
\indent

Assume that $d=d_X=d_Y$ for normalized random vectors $X$ and $Y$. 
Consider the reformulation of the generalized eigenvalue problem in \eqref{sols_pcca}:
\begin{align} \label{sols_pcca_app} 
&\begin{pmatrix}   \Sigma_{X,X\vert Z} & \Sigma_{X,Y\vert Z} \\   
\Sigma_{Y,X\vert Z} & \Sigma_{Y,Y\vert Z} \end{pmatrix}
\begin{pmatrix}   \alpha \\   \beta  \end{pmatrix}= (1+\rho^{(\pcca)}_{X,Y\vert Z})
\begin{pmatrix}   \Sigma_{X,X\vert Z} & 0\\   
0 & \Sigma_{Y,Y\vert Z} \end{pmatrix}
\begin{pmatrix}   \alpha \\   \beta  \end{pmatrix} 
\end{align}
with eigenvalues $\{1+\rho^{(\pcca)}_{X,Y\vert Z}(1),1-\rho^{(\pcca)}_{X,Y\vert Z}(1), \ldots, 1+\rho^{(\pcca)}_{X,Y\vert Z}(d),
1-\rho^{(\pcca)}_{X,Y\vert Z}(d) , 1 ,\ldots ,1 \}$, and where:
\begin{align}\nonumber
\Sigma_{X,Y\vert Z}=\langle X, Y\rangle - \langle X, Z\rangle \langle Z,Z\rangle^{-1} \langle Z,Y\rangle 
\end{align}
For an invertible matrix $D$, the eigenvalues of the generalized eigenvalue problems $Cx=(1+\rho) Dx$ and $D^{-1}Cx=(1+\rho)x$ are identical.
As a result, $\mathcal{C}_{Y \rightarrow X \vert Z }$ can be expressed as:
\begin{align}
\mathcal{C}_{Y \rightarrow X \vert Z } 
&= -\frac{1}{2} \ln \prod_{i=1}^{d} \left[ 1 - \rho_{X,Y \vert Z}^{(\pcca)} (i) ^2 \right] \nonumber \\
&= -\frac{1}{2} \ln \prod_{i=1}^{d} \left[ 1 - \rho_{X,Y \vert Z}^{(\pcca)} (i) \right] \left[ 1 + \rho_{X,Y \vert Z}^{(\pcca)} (i) \right] \nonumber
\end{align}
\begin{align}
&= -\frac{1}{2} \ln   \left| \begin{pmatrix}\Sigma_{X,X\vert Z} & 0\\0 & \Sigma_{Y,Y\vert Z} \end{pmatrix}^{-1} 
\begin{pmatrix}   \Sigma_{X,X\vert Z} & \Sigma_{X,Y\vert Z} \\ \Sigma_{Y,X\vert Z} & \Sigma_{Y,Y\vert Z} \end{pmatrix} \right|  \nonumber \\
&= -\frac{1}{2} \ln \left( \left| \begin{pmatrix}\Sigma_{X,X\vert Z} & 0\\0 & \Sigma_{Y,Y\vert Z} \end{pmatrix} \right| ^{-1} 
\left| \begin{pmatrix}   \Sigma_{X,X\vert Z} & \Sigma_{X,Y\vert Z} \\ \Sigma_{Y,X\vert Z} & \Sigma_{Y,Y\vert Z} \end{pmatrix} \right| \right) \nonumber \\
&= \frac{1}{2} \ln  \left| \Sigma_{X,X\vert Z} \right|  + \frac{1}{2} \ln  \left| \Sigma_{Y,Y\vert Z} \right| 
-\frac{1}{2} \ln \left| \begin{pmatrix}\Sigma_{X,X\vert Z} & \Sigma_{X,Y\vert Z} \\ \Sigma_{Y,X\vert Z} & \Sigma_{Y,Y\vert Z}\end{pmatrix}\right|\nonumber\\
&=H(X\vert Z)+H(Y\vert Z)-H(X,Y\vert Z) \nonumber \\
&=I(X,Y\vert Z) =\mathcal{T}_{Y \rightarrow X \vert Z }. 
\end{align}
Indeed, the transfer entropy $\mathcal{T}_{Y \rightarrow X \vert Z }$ is identical to 
conditional mutual information and the Jensen-Shannon divergence \cite{seg12,mjm}.
Due to \cite{barnett}, and under the specified conditions, it also follows that:
\begin{align}
\mathcal{C}_{Y \rightarrow X \vert Z } &=\frac{1}{2} \mathcal{G}_{Y \rightarrow X \vert Z } \nonumber \\
&= \frac{1}{2} \ln \left( \frac{\left| \Sigma_{X \vert Z} \right| }{\left| \Sigma_{X \vert Y Z} \right| }\right) \label{gewekegenvar}
\end{align}
where the quantity on the right hand side, the generalized variance test, corresponds to the test statistic proposed by Geweke in \cite{geweke82}. 
When $d_X \neq d_Y$ the following relationship is obtained \cite{KICA}:
\begin{align} \nonumber
-\frac{1}{2}\ln \left[ 1 - \max\left(\rho_{X,Y \vert Z}^{(\pcca)}\right)^2 \right]  \leqslant 
\mathcal{T}_{Y \rightarrow X \vert Z } \leqslant 
-\frac{\min(d_X,d_Y) }{2}\ln \left[ 1 - \max\left(\rho_{X,Y \vert Z}^{(\pcca)}\right)^2 \right]
\end{align}
where we have only used the maximum canonical correlation.

\subsection{Link between KCC and transfer entropy}\label{KCWGCTE}
\indent

The link between KCC and transfer entropy stems from the link between kernel generalized variance and mutual information.
Kernel generalized variance approaches a limit as the kernel width approaches zero. 
In the bivariate case, this limit is equal to the mutual information, up to second order, expanding around independence.
A sketch of a proof underlining this property is presented in \cite{KICA}. 
This sketch is based on three main components that can easily be extended to establish a link between KCC and transfer entropy. 
To avoid redundancy, we refer the interested reader to the thorough sketch presented in the mentioned study.

\bibliographystyle{unsrt}
\bibliography{bib.bib}

\end{document}